\documentclass[preprint,review,12pt]{elsarticle}

\usepackage{amssymb}
\usepackage{amsmath}
\usepackage{siunitx}
\usepackage{subcaption} 

\journal{Artificial Intelligence in Agriculture}

\begin{document}

\begin{frontmatter}



\title{Analysis of Plant Nutrient Deficiencies Using Multi-Spectral Imaging and Optimized Segmentation Model}


\author[tll,smart]{Ji-Yan Wu}
\author[tll]{Zheng Yong Poh}
\author[tll]{Anoop C. Patil}
\author[tll]{Bongsoo Park}
\author[smart,goth1,goth2]{Giovanni~Volpe}
\author[tll,nus]{Daisuke Urano}


\affiliation[tll]{organization={Temasek Life Sciences Laboratory}, country={Singapore}}
\affiliation[smart]{Disruptive \& Sustainable Technologies for Agricultural Precision, Singapore-MIT Alliance for Research and Technology (SMART), Singapore}
\affiliation[nus]{organization={Department of Biological Sciences, National University of Singapore}, country={Singapore}}
\affiliation[goth1]{organization={Department of Physics, University of Gothenburg}, country={Sweden}}
\affiliation[goth2]{organization={Science for Life Laboratory, Department of Physics, University of Gothenburg}, country={Sweden}}

\begin{abstract}
Accurate detection of nutrient deficiency in plant leaves is essential for precision agriculture, enabling early intervention in fertilization, disease, and stress management. This study presents a deep learning framework for leaf anomaly segmentation using multispectral imaging and an enhanced YOLOv5 model with a transformer-based attention head. The model is tailored for processing nine-channel multispectral input and uses self-attention mechanisms to better capture subtle, spatially-distributed symptoms. The plants in the experiments were grown under controlled nutrient stress conditions for evaluation. We carry out extensive experiments to benchmark the proposed model against the baseline YOLOv5. Extensive experiments show that the proposed model significantly outperforms the baseline YOLOv5, with an average Dice score and IoU (Intersection over Union) improvement of about $12\%$. In particular, this model is effective in detecting challenging symptoms like chlorosis and pigment accumulation. These results highlight the promise of combining multi-spectral imaging with spectral-spatial feature learning for advancing plant phenotyping and precision agriculture.
\end{abstract}




\end{frontmatter}



\section{Introduction}
The strategic implementation of precision agriculture technologies, particularly image-based anomaly detection, is transforming modern agricultural practices~\cite{talaviya2024aiag, Zaman-Allah2015Multispectral}. With the help of satellite, drone, and proximal imaging, farmers can now monitor crop health at scale to optimize fertilization, detect early signs of nutrient deficiency, and manage disease outbreaks~\cite{Papachristoforou2023Multispectal}. However, to make these systems truly automated and actionable, robust computer vision algorithms capable of accurately detecting anomalies across thousands of field images are needed.

Most current image-based plant health monitoring systems rely on conventional RGB cameras, which are limited to three spectral bands. In contrast, multi-spectral imaging captures reflectance data across a broader spectral range—including near-infrared (NIR), and even short wave-length infrared bands (SWIR)—which provide richer information on plant physiology and stress response~\cite{Claudio2021Multispectral, Charles2019Multispectral}. NIR reflectance, in particular, is closely associated with leaf senescence and chlorophyll content, and has been widely used in vegetation indices such as Normalized Difference Vegetation Iindex (NDVI) and Normalized Difference Red-Edge index (NDRE)~\cite{RuiXu2019Multispectal}.
Affordable multi-spectral cameras typically use dual-sensor configurations to separately capture visible and NIR bands, which is advantageous for aerial imaging but introduces parallax and alignment issues at close range~\cite{Boris2022Multispectral}. In contrast, multi-channel sensors with a shared optical path eliminate registration errors and enable precise analysis of plant features from the same viewpoint. 

Deep learning methods have shown strong promise in plant anomaly detection tasks~\cite{Peng2022Multispectral}, outperforming traditional machine learning models in both accuracy and adaptability. However, most state-of-the-art deep learning architectures, such as YOLO and Mask R-CNN, are designed for RGB images and must be adapted to handle high-dimensional multi-spectral data. Challenges include input channel compatibility, feature fusion across spectral bands, and generalization under varying stress scenarios. Recent studies have also shown the effectiveness of combining multi-spectral sensing with AI-based models for early plant stress detection. For instance, Wei et al.~\cite{wei2023forest} used multi- and hyper-spectral images to classify disease symptoms in forests. Similarly, deep learning models have been applied for plant phenotyping and genebank analysis across multi-spectral inputs~\cite{zhang2024plantai}.

This study addresses the following key challenges in applying deep learning to multi-spectral leaf analysis.
On the one hand, there is a need for spectral feature selection and fusion: Given the increased number of spectral bands, it is critical to design models that can selectively emphasize the most informative wavelengths for detecting early-stage anomalies.
On the other hand, also the model architecture needs to be adapted: Conventional deep learning models require modifications to process 9-channel inputs and perform pixel-wise segmentation across subtle, overlapping symptom classes.

To address these challenges, we propose a deep learning framework based on the YOLOv5 object segmentation architecture~\cite{jocher2021ultralytics}, enhanced with a transformer-based attention head~\cite{vaswani2017attention} to improve contextual reasoning across spectral-spatial features. Our model is trained on a newly constructed dataset of \textit{Marchantia polymorpha} grown under five controlled nutrient conditions: fully fertilized control as well as deficiencies in nitrate, phosphate, calcium, and iron.

A nine-channel multi-spectral camera was used to capture reflectance data over 17 days. Pixel-wise annotations were made for four visually distinct symptom categories: normal, chlorosis, pigment accumulation, and tipburn. These symptom classes correspond directly to specific nutrient deficiencies observed in our controlled experiments. In particular, chlorosis (yellowing of leaf tissue) is commonly associated with iron and nitrate deficiencies, while pigment accumulation (typically reddish or purple discoloration) results from stress due to phosphate or nitrate imbalance. Tipburn, characterized by necrotic leaf margins, is a hallmark of calcium deficiency. The normal class includes uniformly healthy green tissue grown under fully fertilized conditions. By establishing this clear linkage between visual symptoms and physiological nutrient stressors, our annotation framework enables biologically meaningful segmentation for downstream phenotyping and stress classification. Experimental results demonstrate that the proposed method outperforms the RGB-only YOLOv5 baseline, especially in detecting small or scattered lesion patterns.

The main contributions of this work are the following.
First, we develop a novel leaf anomaly segmentation framework that extends the YOLOv5 architecture to support nine-channel multi-spectral image input. To better capture subtle spatial and spectral variations, we integrate a transformer-based attention head into the YOLOv5 output layers. This optimization enhances the model’s ability to detect small or dispersed defects such as chlorosis and marginal pigment accumulation, which are often underrepresented in RGB-based detection models.
Second, we construct a high-quality, pixel-level annotated dataset of multi-spectral plant images under controlled nutrient stress conditions. The dataset spans five nutrient regimes (control, -Fe, -N, -P, -Ca) and includes visual and spectral symptoms across three plant species. Each image is annotated with fine-grained polygon masks representing four classes (normal, chlorosis, pigment accumulation, tipburn), enabling robust supervised segmentation training and evaluation.
Finally, we conduct extensive experiments comparing our proposed model against the baseline RGB-based YOLOv5. The results demonstrate consistent improvements in segmentation performance, achieving up to 17\% higher mean Average Precision (mAP) and 12\% higher average Dice score across all classes. These findings confirm the effectiveness of integrating multi-spectral information and transformer-based attention for plant stress detection.

The rest of this article is organized as follows: Section~\ref{Sec2} reviews related work on multi-spectral imaging and object segmentation. Section~3 introduces the architecture and training strategy. Section~4 presents the experimental design, dataset, and evaluation results. Section~5 concludes the study and outlines future research directions.

\section{Related Work}
\label{Sec2}
Relevant previous works can be divided into three categories: i) the analysis of plant healthy and disease status using multi-spectral imaging; ii) deep learning models for object detection and segmentation; iii) deep learning models for multi-spectral and hyperspectral image analysis. The following subsections discuss the previous efforts in these areas.

\subsection{Plant Healthy Analysis Using Multi-Spectral Images}
RGB imaging with three wavelength channels has been the most prevalent data for plant anomaly detection. Recently, multispectral and hyperspectral imaging are gaining traction for their higher spectral resolution and sensitivity to plant health~\cite{Li2022HyperspectralReview}.

One notable study was conducted by Charles Veys et al.~\cite{Charles2019Multispectral} who demonstrated  proposed usingthe use of multispectral imaging as a non-invasive method diagnostics for detecting presymptomatic X phenotype for in oilseed rape. A one-class support vector machine (SVM) was used for the healthy- and stressed-plant classificationsthis study, achieving a $92\%$ accuracy. Similarly, Yao Peng et al. ~\cite{Peng2022Multispectral} followed a similar approach by incorporatingemployed spatial-spectral machine learning for the early detection of plant virus infectiones, achieving over $85\%$ classification accuracy. Additionally, Claudio I. Fernández et al. ~\cite{Claudio2021Multispectral} utilized multispectral imagingery to detect infections in for infected cucumbers. They  plant detection, achieved ing $89\%$ accuracy with using RGB-derived features alone, however when although combining all 5 spectral bands (Blue (475 nm), Green (560 nm), Red (668 nm), Red-edge (717 nm), were Near-Infrared (NIR) (840 nm)) combined, the accuracy dropped to reduced the performance to $57\%$., suggesting possible issues with feature integration processes. 

Comparative studies have explored the integration of ing different imaging sensing techniques, Boris Lazarevic et al. ~\cite{Boris2022Multispectral} conducted multispectral imaging, chlorophyll fluorescence, and 3D scanning for the detection of nutrient deficienciesy detection in common beans. Using Recursive recursive partitioning, they achieved effectively differentiated between effective phenotypes segregation.

At a larger scale, aerial multispectral sensors with Unmanned Aerial Platforms (UAPs) have been used for field analysis, nutrient deficiency identification~\cite{Zaman-Allah2015Multispectral,RuiXu2019Multispectal}, ecosystem monitoring~\cite{Wolff2023Multispectal}, and plant classification~\cite{Papachristoforou2023Multispectal}. For example, Wolff Franziska and coworkers~\cite{Wolff2023Multispectal} reported accuracies of 0.59-0.82 in mapping plant communities using Random Forest classifiers.
More recently, Nguyen et al.~\cite{Nguyen2022Hyperspectral} applied spectral vegetation indices from multispectral UAV images to monitor maize nitrogen deficiency. Their results showed that narrow-band multispectral data significantly improves the accuracy over conventional RGB approaches.

\subsection{Deep Learning Models for Object Detection and Segmentation}

Deep learning has significantly advanced object detection and segmentation in computer vision. Modern object detection models are generally categorized into two paradigms: two-stage detectors and one-stage detectors. Two-stage detectors such as Faster R-CNN~\cite{ren2015faster} first generate region proposals and then perform classification, while one-stage detectors like YOLO~\cite{redmon2018yolov3} and SSD~\cite{liu2016ssd} directly predict bounding boxes and classes in a single pass.

YOLOv3~\cite{redmon2018yolov3} introduced a multi-scale detection strategy using Darknet-53 backbone and remains widely used due to its speed and accuracy. Later versions such as YOLOv4~\cite{bochkovskiy2020yolov4} and YOLOv5~\cite{jocher2021yolov5} incorporated Cross Stage Partial (CSP) networks, spatial pyramid pooling (SPP), and path aggregation networks (PANet) to improve efficiency. YOLOv7 and YOLOv8 further optimized architectural depth and used decoupled head designs to improve training convergence and performance.

Several segmentation models such as Mask R-CNN~\cite{he2017mask} extend Faster R-CNN by adding a mask prediction branch. DeepLabV3+~\cite{chen2018encoder} employs Atrous Spatial Pyramid Pooling (ASPP) and encoder-decoder structure to capture fine details in semantic segmentation. More recently, Segment Anything~\cite{kirillov2023segment} has shown impressive generalization capabilities using vision transformers trained with large-scale data.

Attention mechanisms have become pivotal in improving detection accuracy. Convolutional Block Attention Module (CBAM )~\cite{woo2018cbam} and SE (Squeeze-and-Excitation) networks~\cite{hu2018squeeze} are popular channel/spatial attention blocks used to enhance feature maps. In our work, we incorporate CBAM within the YOLOv5 backbone to better capture subtle features in multi-spectral channels.

Transformer-based detection has also gained traction. DETR~\cite{carion2020detr} replaces traditional region proposal networks with transformer encoders and decoders. Deformable DETR~\cite{zhu2021deformable} improves DETR’s convergence and accuracy by attending to sparse key points in feature maps. These studies inspire our design of a transformer-based head in YOLOv5 to improve segmentation of small and scattered leaf anomalies.

Few works have extended these architectures to non-RGB data. Notably, Nataprawira et al.~\cite{nataprawira2021pedestrian} used YOLOv3 on multi-spectral pedestrian data, but without a segmentation focus. Our method, in contrast, tailors YOLOv5 with a transformer head and attention modules for fine-grained segmentation on multi-spectral plant leaf images, addressing challenges specific for agricultural imaging.

\subsection{Deep Learning on Multi-Spectral and Hyperspectral Image Analysis}

While RGB-based object detection is well established, adapting deep learning models to multi-spectral and hyperspectral data poses unique challenges, such as high-dimensional inputs and spectral feature fusion. Nataprawira et al.~\cite{nataprawira2021pedestrian} used YOLOv3 for pedestrian detection with multi-spectral images, demonstrating superior detection rates at night compared to RGB-only inputs. Ma et al.~\cite{ma2021hyperspectral} proposed a hyperspectral object detection model based on Faster-RCNN, integrating spectral-spatial attention mechanisms. Their method achieved significant improvements in precision compared to models treating spectral bands independently. Zhao et al.~\cite{zhao2022multispectral} introduced an attention-guided multi-branch CNN for pest detection in multi-spectral images, showing the importance of channel-wise attention in spectral domain. Li et al.~\cite{li2023adaptive} proposed an adaptive spectral-spatial feature fusion network (ASSFN) for hyperspectral disease detection in crops. Their results demonstrated that proper spectral feature aggregation greatly boosts plant disease classification accuracy. Finally, Zhang et al.~\cite{zhang2023hyperspectraltransformer} developed a transformer-based model for hyperspectral image classification, illustrating the superior ability of transformers to model long-range spectral dependencies compared to CNN-based models.

The above studies on multi- and hyperspectral image analysis suggest that the combination of attention mechanisms, spectral fusion, and transformer architectures is promising for future research in multi-spectral plant disease detection, as explored in our work.

The closest studies to this paper are on the object detection using multi-spectral images (e.g., \cite{nataprawira2021pedestrian}). The authors in \cite{nataprawira2021pedestrian} use YOLOv3 to detect pedestrians with multi-spectral images. To the best of our knowledge, this is the first work on multi-spectral image analysis using deep learning detection model for plant healthy diagnosis.

\section{Proposed Method}
\label{sec:method}
This section introduces the two most important aspects of the proposed method: i) neural network architecture for multi-spectral image segmentation; ii) multi-spectral image data-set collection and preparation. The above two functions/steps are essential to conduct the plant nutrition deficiency analysis using multi-spectral imaging data.

\subsection{Neural Network Architecture}

\begin{figure}[b!]
\centering
\includegraphics[width=1\linewidth,keepaspectratio]{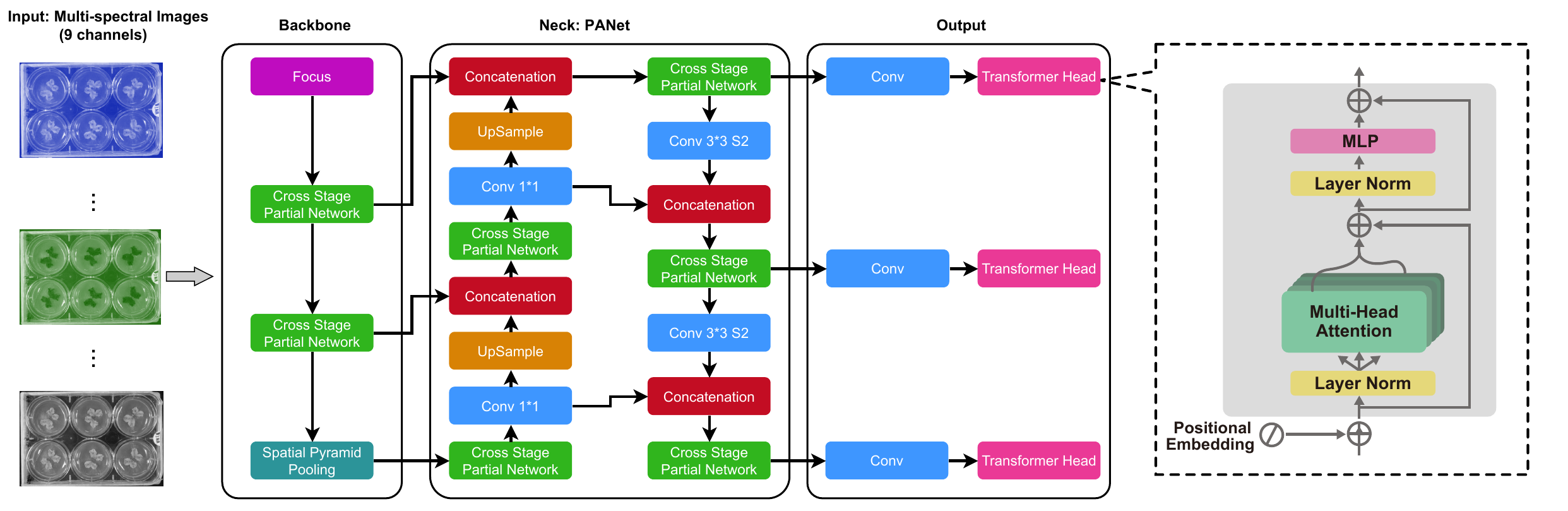}
\caption{Overview of the neural network architecture for multi-spectral image analysis. Nine-channel multi-spectral images ($640 \times 640$ pixels) are passed to YOLOv5, consisting of three main modules: Backbone (CSPNet + SPP), Neck (PANet), and Output Head (Transformer-based segmentation layers). 
As shown on the right, we modified the head to incorporate transformer encoder blocks for fine-grained pixel-wise prediction, tailored to small-size plant anomalies.}
\label{fig:design}
\end{figure}

The proposed architecture is an enhancement of YOLOv5~\cite{jocher2021ultralytics}, modified to accommodate and fully exploit nine-channel multi-spectral input images. As shown in Fig.~\ref{fig:design}, the architecture consists of three major modules: a CSP-based Backbone for feature extraction, a PANet Neck for multi-scale fusion, and a Transformer Head for semantic segmentation. Below, we elaborate on each component and the specific adaptations introduced.

\textbf{Design Motivation.}
Through the combination of efficient CSP encoding, multi-scale PANet fusion, and a global context-aware transformer head, the network is tailored to segment small and irregular leaf anomalies from nine-channel multi-spectral data. This hybrid design balances computational efficiency with spectral sensitivity, addressing the challenges in multi-spectral plant phenotyping.

\textbf{1. Multi-Channel Input Adaptation.}
To process multi-spectral images, we extend the YOLOv5 architecture to accept 9 input channels instead of the default 3 (RGB). This modification is applied at the very beginning of the network—specifically in the first convolutional layer of the Focus module. In the original YOLOv5, the Focus layer expects an input tensor of shape $(3, H, W)$. We modify this to accept tensors of shape $(9, H, W)$ by adjusting the number of input channels in the corresponding convolution kernel from 3 to 9.

To leverage pretrained weights from the standard RGB model while supporting 9-channel input, we apply the following adaptation strategy: the original 3-channel weights are copied and replicated to initialize the first three channels of the new convolutional kernel, and the remaining six channels are initialized either randomly (e.g., using Kaiming initialization) or by averaging the original weights. This hybrid initialization enables the model to retain useful low-level feature representations from the pretrained model while allowing flexibility to learn spectral-specific features during fine-tuning.

The Focus module itself operates as in the standard YOLOv5: it slices and rearranges spatial information into channel-wise representations to enrich feature diversity. This adaptation ensures that all nine spectral channels contribute to early-stage feature extraction and that their distinct spectral signatures are preserved throughout the network.

The standard YOLO dataloader does not support variable input channels; hence, we introduced a customized dataloader to correctly normalize and batch 9-channel inputs, ensuring consistent spectral ordering across training and inference. The standard YOLOv5 dataloader is designed for 3-channel RGB inputs and does not natively support multi-spectral images with arbitrary channel counts. To accommodate 9-channel input tensors, we implemented a customized dataloader that performs proper normalization, batching, and spectral channel ordering. It ensures that the input bands are consistently arranged across training and inference steps (e.g., channel 1 always corresponds to 470 nm, channel 2 to 530 nm, etc.).

Maintaining this fixed spectral order is critical because each spectral band captures reflectance properties at a specific wavelength, and these wavelengths exhibit unique responses to plant physiology under nutrient stress. For example, near-infrared bands are highly sensitive to internal leaf structure and water content, while visible bands (e.g., blue or red) respond to pigment changes like chlorosis or anthocyanin accumulation. If band ordering is inconsistent or misaligned across samples, the model may learn spurious correlations and fail to associate specific wavelengths with their biological signatures. Thus, spectral alignment during data loading is a prerequisite for reliable feature fusion and effective learning in multi-spectral image analysis.

\textbf{2. CSP-Based Backbone for Efficient Feature Encoding.}
The backbone utilizes a Cross Stage Partial Network (CSPNet) with several advantages: reduced computational cost, enhanced gradient propagation, and lower memory usage. Each CSP block splits feature maps into two branches: one undergoing heavy computation and the other acting as a skip connection. These are later merged, enabling a balance between depth and efficiency.

The backbone also incorporates Spatial Pyramid Pooling (SPP) to extract multi-scale contextual information through parallel pooling operations of varying kernel sizes. SPP improves the network's ability to detect defects of varying sizes and shapes across the plant surface.

\textbf{3. PANet Neck for Multi-Scale Feature Fusion.}
The Path Aggregation Network (PANet) acts as the Neck of the architecture, propagating multi-resolution features from the backbone to the head. It performs successive upsampling, downsampling, and lateral connections via concatenation operations. Each stage fuses low-level spatial detail with high-level semantic information.

This multi-scale fusion mechanism is particularly useful in our setting where plant defects (e.g., tipburn, pigment accumulation) vary in size, shape, and location. The use of 1$\times$1 and 3$\times$3 convolution blocks after each concatenation helps refine channel-wise interaction and spatial alignment.

\textbf{4. Transformer Head for Fine-Grained Segmentation.}
To address the limitation of standard YOLO heads in detecting fine-grained, small-scale defects, we propose a Transformer-based segmentation head inspired by vision transformers~\cite{vaswani2017attention}. Unlike traditional heads that rely purely on convolutional operations, the transformer head captures global spatial-spectral dependencies through self-attention.

Each transformer encoder consists of the following components:
\begin{itemize}
    \item Positional Embedding: Added to maintain spatial order, since attention mechanisms are permutation-invariant.
    \item Multi-Head Self-Attention (MHSA):
    \begin{equation}
        \text{Attention}(Q, K, V) = \text{softmax}\left( \frac{QK^\top}{\sqrt{d_k}} \right)V
    \end{equation}
    where $Q$, $K$, and $V$ are query, key, and value projections of the input features.
    \item Residual Connections \& Layer Normalization: Applied before and after attention and Multi-Layer Perceptron (MLP) blocks for training stability.
    \item Feed-Forward Network: A two-layer fully connected network with GELU activation.
\end{itemize}

The final output of the transformer head is a high-resolution segmentation map, predicting defect regions at the pixel level. Using the transformer head after Convolutional layers helps retain global coherence while improving the localization of scattered and low-contrast anomalies.

\begin{table}[htbp]
\renewcommand{\arraystretch}{1.25}
\scriptsize
\centering
\caption{Conceptual comparison between the baseline YOLOv5 and our proposed YOLOv5 with transformer head for multi-spectral leaf defect segmentation.}
\label{tab:yolo_comparison}
\begin{tabular}{|>{\raggedright\arraybackslash}m{4.2cm}|>{\raggedright\arraybackslash}m{4.4cm}|>{\raggedright\arraybackslash}m{4.4cm}|}
\hline
\textbf{Component} & \textbf{Baseline YOLOv5} & \textbf{Proposed Model} \\
\hline
\textbf{Input Format} & RGB (3 channels) & Multi-spectral (9 channels) \\
\hline
\textbf{Data Loader Support} & Standard PyTorch dataloader with fixed 3-channel input & Customized dataloader to accommodate 9-channel spectral data with aligned pre-processing \\
\hline
\textbf{Backbone Architecture} & CSPNet with Focus and SPP modules & Same, but extended to process 9-channel data from the first layer onward \\
\hline
\textbf{Detection/Segmentation Head} & Convolutional prediction head for detection or segmentation & Transformer encoder block for pixel-wise semantic segmentation using self-attention \\
\hline
\textbf{Attention Mechanism} & Not included by default & Vision Transformer block with MHSA, LayerNorm, and residual connections \\
\hline
\textbf{Small Defect Sensitivity} & Moderate; performance degrades on tiny targets & Enhanced by transformer head focusing on global and contextual information \\
\hline
\end{tabular}
\end{table}

\noindent

\textbf{Comparison with Baseline YOLOv5.} As shown in Table~\ref{tab:yolo_comparison}, our modifications extend the baseline YOLOv5 framework in multiple directions. By introducing support for 9-channel input and replacing the standard prediction head with a transformer-based encoder, we significantly enhance the model’s capability to identify subtle, small-sized defects in complex spectral environments. This is particularly important in agricultural applications where symptoms such as chlorosis or tipburn may occupy only a small portion of the leaf surface, and contextual spectral cues are essential for detection. These architectural adjustments lay the foundation for the improved segmentation performance observed in later evaluation.



\subsection{Dataset Collection and Preparation}

\textbf{Controlled Nutrient Stress Experiment.}
To generate a reliable dataset for leaf anomaly detection, we conducted a nutrient-controlled plant growth experiment using \textit{Marchantia polymorpha}. Each plant was grown in a 6-well petri dish under one of five nutrition conditions: a fully fertilized control and four nutrient-deficient treatments corresponding to $0\%$ iron (Fe), $3\%$ nitrate (N), $3\%$ phosphate (P), and $3\%$ calcium (Ca), each inducing specific phenotypic stress responses over time.

\begin{figure}[htbp]
\centering
\includegraphics[width=\linewidth]{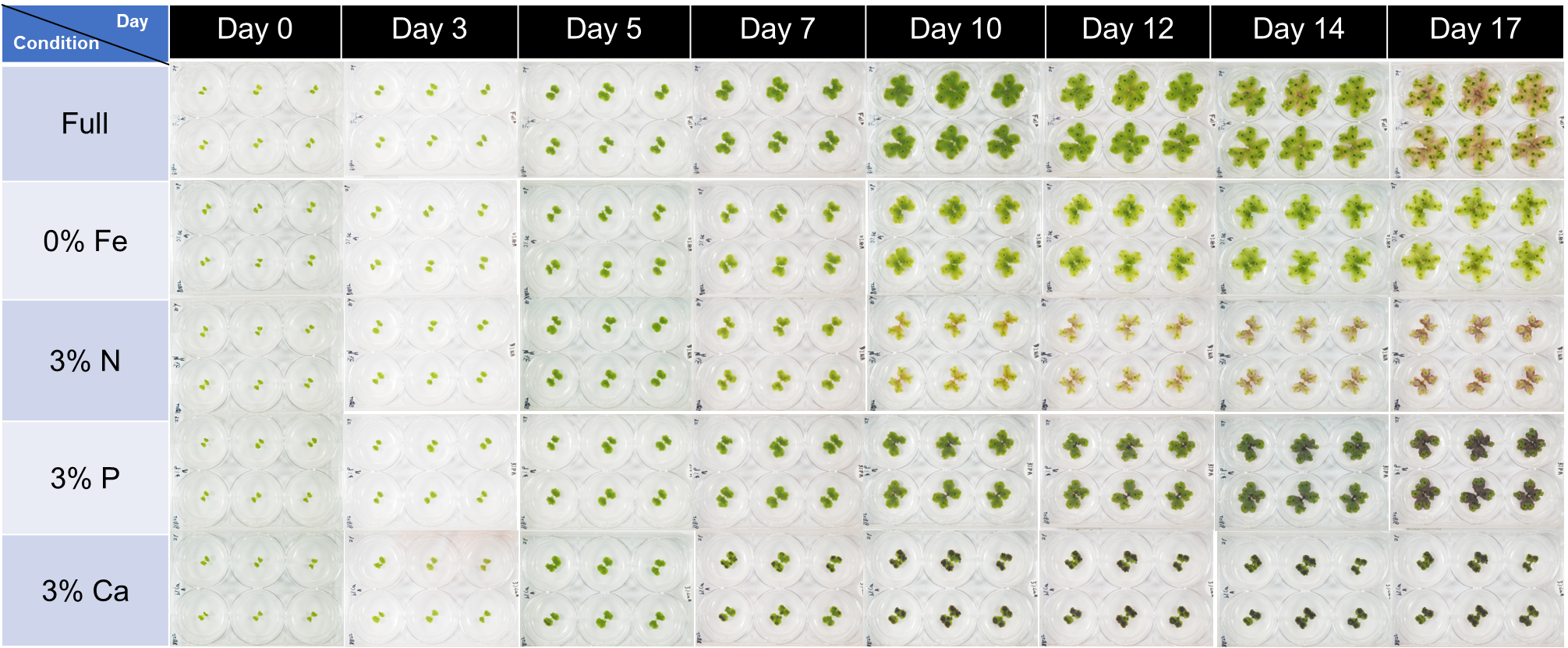}
\caption{Representative RGB images from the nutrient stress experiment over 17 days on Marchantia. Each row corresponds to a specific nutrient condition, with columns indicating time points. Morphological and color differences due to stress are visually apparent.}
\label{fig:experiment_timeline}
\end{figure}

Multi-spectral and RGB images were captured from top-down view at eight time points: Days 0, 3, 5, 7, 10, 12, 14, and 17. Imaging was performed under controlled lighting with consistent background settings. The lights are calibrated to maintain consistent irradiance across the field of view and avoid spectral bias toward any particular wavelength region. Additionally, the background surface was standardized using a non-reflective matte black material to enhance contrast between leaf edges and surrounding pixels. These conditions ensure consistent reflectance measurements across the 9 spectral bands and minimized variability unrelated to plant phenotypes, thereby improving the signal quality for stress-related feature extraction.Fig.~\ref{fig:experiment_timeline} illustrates representative RGB images of plants under each condition across the 17-day experiment timeline.

\textbf{Phenotypic Observations.}
As shown in Fig.~\ref{fig:experiment_timeline}, distinct growth patterns and visual symptoms emerged under different nutrient deficiencies: i) Full nutrition (control): Plants showed healthy, uniform growth with consistent dark green coloration and circular morphology by Day 17. ii) $0\%$ Fe (Iron deficiency): Chlorosis was evident as early as Day 5, progressing to severe yellowing across leaf tissue by Day 10. Overall plant size was reduced. iii) $3\%$ N (Nitrate deficiency): Plants remained pale green from Day 3 onward, and pigment accumulation appeared along leaf margins by Day 10. By Day 14, growth stunted and necrotic patches formed. iv) $3\%$ P (Phosphate deficiency): Dark purple pigment accumulation was visible near central lobes starting from Day 10, intensifying by Day 17. Leaf expansion was relatively preserved compared to other treatments. v) $3\%$ Ca (Calcium deficiency): Tipburn and marginal necrosis became visible around Day 7–10, with leaves curling and darkening by Day 14. Necrotic areas spread irregularly toward leaf centers.

The color and shape differences across conditions make them suitable for both human-visual assessment and AI-driven anomaly detection.

\begin{figure}[htbp]
\centering
\includegraphics[width=1\linewidth]{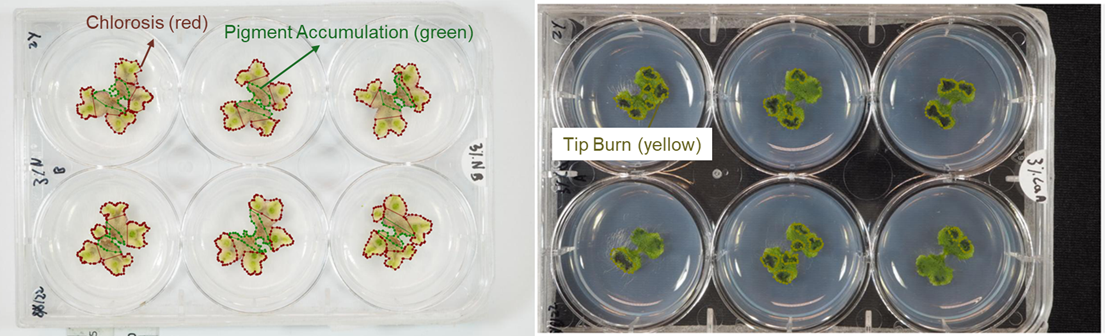}
\caption{Left: Annotated images with polygon masks for chlorosis, pigment accumulation, and tipburn. Right: Corresponding original multi-spectral image plates used as input to the segmentation model.}
\label{fig:annotation_examples}
\end{figure}
\textbf{Annotation and Labeling Procedure.}
All collected images were manually annotated using the open-source LabelMe \cite{labelme} tool. Annotations were performed using polygon masks to delineate distinct defect regions. Four classes were labeled: normal tissue, chlorosis, pigment accumulation, and tipburn. Each defect type was annotated with a separate polygon, ensuring tight boundaries around the affected areas. The annotation guidelines included i) Chlorosis: defined as uniformly yellowed areas with reduced pigment intensity; ii) Pigment Accumulation: deep purple or reddish patches, often along veins or leaf centers; iii) Tipburn: blackened, necrotic regions starting at the leaf edge or tips; iv) Normal: healthy, green regions without discoloration.

Fig.~\ref{fig:annotation_examples} illustrates representative annotated images of plants with N deficiency or Ca deficiency defects. The left panel shows polygon annotations overlaid on visible light images; the right panel shows original multi-spectral captures for the same samples.
All images were resized to $640 \times 640$ pixels and normalized before training. The dataset was split into $90\%$ training and $10\%$ validation sets. Annotation consistency was maintained through double-checking by two reviewers and cross-validation on ambiguous samples.
This carefully annotated dataset provides a robust foundation for training pixel-wise segmentation models under realistic plant stress scenarios.

\section{Experiments}
This section presents the experimental setup and results to evaluate the effectiveness of the proposed multi-spectral leaf defect segmentation model. As described in Section~\ref{sec:method}, multi-spectral image datasets were collected from Marchantia polymorpha grown under controlled nutrient conditions. The dataset includes nine-channel multi-spectral images annotated with pixel-wise labels corresponding to four symptom classes: normal tissue, chlorosis, pigment accumulation, and tipburn as shown in Fig.\ref{fig:annotation_examples}.

The experiments are designed to assess the model’s segmentation performance, robustness across symptom types, and sensitivity to small and dispersed defect regions. We detail the training parameters, evaluation metrics, and provide both quantitative results and qualitative visualizations to validate the model’s segmentation capabilities.

\subsection{Experimental Setting}
This section presents the evaluation setup and experimental results on validating the performance of our optimized segmentation model using annotated multi-spectral data-set. The first subsection presents the experimental settings. Then, we present and discuss the evaluation results in detail. 
\textbf{Plant assay preparation.}
The liverwort \textit{Marchantia polymorpha} was selected as the primary plant model due to its simple, flat thallus morphology, which allows uniform top-down imaging and consistent reflectance measurement across the surface. Wild-type Tak-1 gemmalings were first cultured on $1/2 \times$ B5 media without sucrose under continuous light ($41.4$–$46$ $\mu$mol photons/m\textsuperscript{2}/s) for 10 days to establish the healthy baseline growth.

Following pre-cultivation, plants with similar size and morphology were transferred into sterile 6-well culture plates for treatment. The experimental design included one control (full nutrition using Yamagami medium) and four nutrient stress conditions: $0\%$ Iron (Fe), $3\%$ Phosphate (P), $3\%$ Nitrate (N), and $3\%$ Calcium (Ca). The nutrient-deficient media were prepared by omitting or reducing specific components from the Yamagami formulation. Each condition had 12 biological replicates to ensure statistical robustness.

Images were acquired under controlled lighting using both RGB and multi-spectral cameras on treatment days 0, 3, 5, 7, 10, 12, 14, and 17. Multi-spectral images consisted of 9 aligned spectral bands ranging from visible to near-infrared wavelengths.

\textbf{Dataset description.}
In summary, the experimental dataset includes the following:
\begin{itemize}
    \item \textbf{Multi-spectral images} (9 channels): Used for training and evaluating the proposed model. These images preserve spectral responses essential for detecting stress-related symptoms.
    \item \textbf{RGB images} (3 channels): Extracted from three selected bands (i.e., $470$ nm, $530$ nm, $620$ nm) to train and evaluate the baseline YOLOv5 model. They are also used for human-readable visualization and overlay purposes.
\end{itemize}

A total of 160 multi-spectral images were manually annotated using the LabelMe tool in polygon mode. Each image contains polygon masks labeled into one of four visually and physiologically distinct classes:
\begin{itemize}
    \item \textbf{Normal}: Uniform green leaf tissue with no visible stress.
    \item \textbf{Chlorosis}: Yellowing or pale areas due to iron/nitrogen deficiency.
    \item \textbf{Pigment Accumulation}: Reddish or purple patches typically caused by phosphate or nitrate imbalance.
    \item \textbf{Tipburn}: Necrotic, darkened lesions, typically near leaf margins caused by calcium deficiency.
\end{itemize}

Each annotated image contains multiple defect regions, and each polygon is class-labeled independently. Annotations were reviewed by two independent annotators to ensure consistency and accuracy. Some images contain overlapping or adjacent symptom types, especially in nutrient-deficient samples.

\begin{figure}[htbp]
\centering
\includegraphics[width=0.495\linewidth]{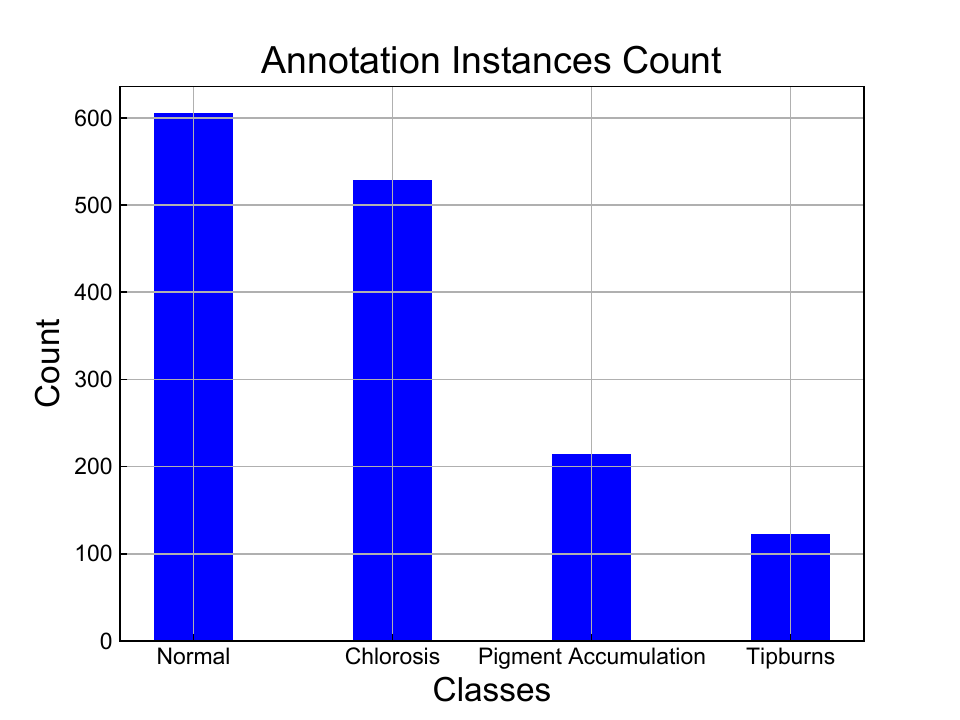}
\includegraphics[width=0.495\linewidth]{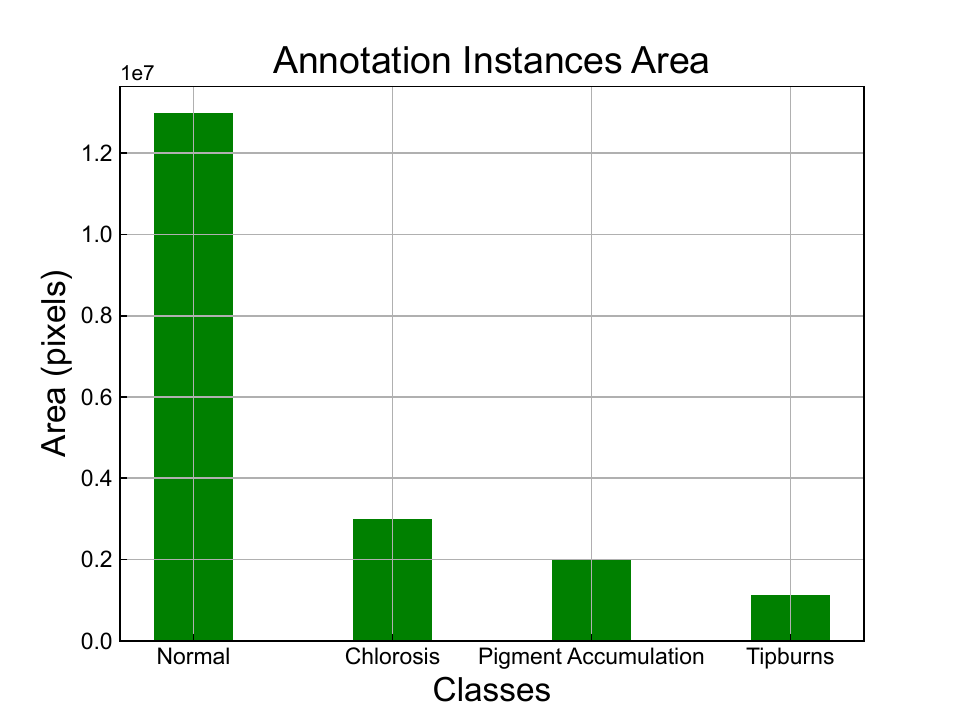}
\caption{Summary of annotated instances in the multi-spectral dataset. (Left) Total number of polygons per class. (Right) Aggregate annotated area (in pixels) per class.}
\label{fig:anno}
\end{figure}

The dataset was randomly split into a 90\% training set and a 10\% validation set to ensure statistical consistency while maintaining representation of all four classes in both sets. Fig.~\ref{fig:anno} summarizes the annotation statistics. Fig.~\ref{fig:anno}a shows the number of annotated instances per class, while Fig.~\ref{fig:anno}b illustrates the cumulative pixel area of each class label across the dataset.

The class distribution reveals that the Normal class has the highest number of annotations and coverage area, while Tipburn and Pigment Accumulation are more localized and sparse. This imbalance presents additional challenges for model training, particularly for rare or small-area symptoms like chlorosis and tipburn, which benefit from spectral enhancement and attention-based detection modules introduced in our proposed model.

\textbf{Performance metrics.}
To comprehensively assess segmentation performance, we employ the following commonly used metrics in semantic segmentation tasks. The first is intersection over union (IoU):
\begin{equation}
        \text{IoU} = \frac{|\text{Prediction} \cap \text{Ground Truth}|}{|\text{Prediction} \cup \text{Ground Truth}|}
\end{equation}
which ,measures the pixel-wise overlap between predicted and ground truth masks.

The second one is thew Dice score (also known as F1-score in segmentation tasks):
\begin{equation}
        \text{Dice} = \frac{2 \cdot |\text{Prediction} \cap \text{Ground Truth}|}{|\text{Prediction}| + |\text{Ground Truth}|}
\end{equation}
which emphasizes correct pixel classification, especially useful for small targets.

We also used precision and recall:
    \begin{equation}
        \text{Precision} = \frac{TP}{TP + FP}, \quad \text{Recall} = \frac{TP}{TP + FN}
\end{equation}
which evaluate the ratio of correctly identified pixels to all predicted/actual pixels.

Finally, we used the Mean Average Precision (mAP), calculated across confidence thresholds and averaged over classes. This reflects the trade-off between precision and recall at multiple decision boundaries.

These metrics are reported both globally (over the whole dataset) and per class to analyze performance on subtle symptoms like tipburn.

\textbf{Training and evaluation settings.}
The model was implemented using the PyTorch framework and trained on a single NVIDIA RTX 3080 GPU. Augmentations included random flipping, rotation, and color jitter. These augmentations are introduced for increasing the variety of plant images under controlled imaging conditions. First, leaf orientation is not fixed in real growth settings—leaves may be positioned at different angles or flipped due to random growth or manual handling. Second, color jittering introduces slight changes in brightness and contrast, which mimics subtle differences in reflectance caused by minor fluctuations in lighting, pigment concentration, or sensor calibration. 

The training hyperparameters are summarized in Table~\ref{tab:train_parameter}. A total of 500 epochs were run using Stochastic Gradient Descent (SGD) with momentum. Batch size was set to 4 due to GPU memory constraints with 9-channel input.

\begin{table}[htbp]
  \centering
  \caption{Training hyperparameters for segmentation model.}
  \small
    \begin{tabular}{|c|c||c|c|}
        \hline
        \textbf{Parameter} & \textbf{Value} & \textbf{Parameter} & \textbf{Value} \\
        \hline
        Epochs & $500$ & Image size & $640 \times 640$ \\
        Learning rate & $1 \times 10^{-4}$ & Momentum & $0.99$ \\
        Optimizer & SGD & Batch size & $4$ \\
        Input channels & $9$ (multi-spectral) & Augmentations & Flip, rotate, color jitter \\
        \hline
    \end{tabular}
    \label{tab:train_parameter}
\end{table}

\subsection{Evaluation Results}

In this section, we evaluate the segmentation performance of the proposed multi-spectral YOLOv5 model with transformer head, and compare it against the baseline YOLOv5 trained on standard RGB images. Both quantitative metrics and qualitative visualizations are provided to demonstrate the model’s effectiveness in capturing nutrient deficiency symptoms at pixel level.

\begin{figure}[htbp]
\centering
\includegraphics[width=0.495\linewidth]{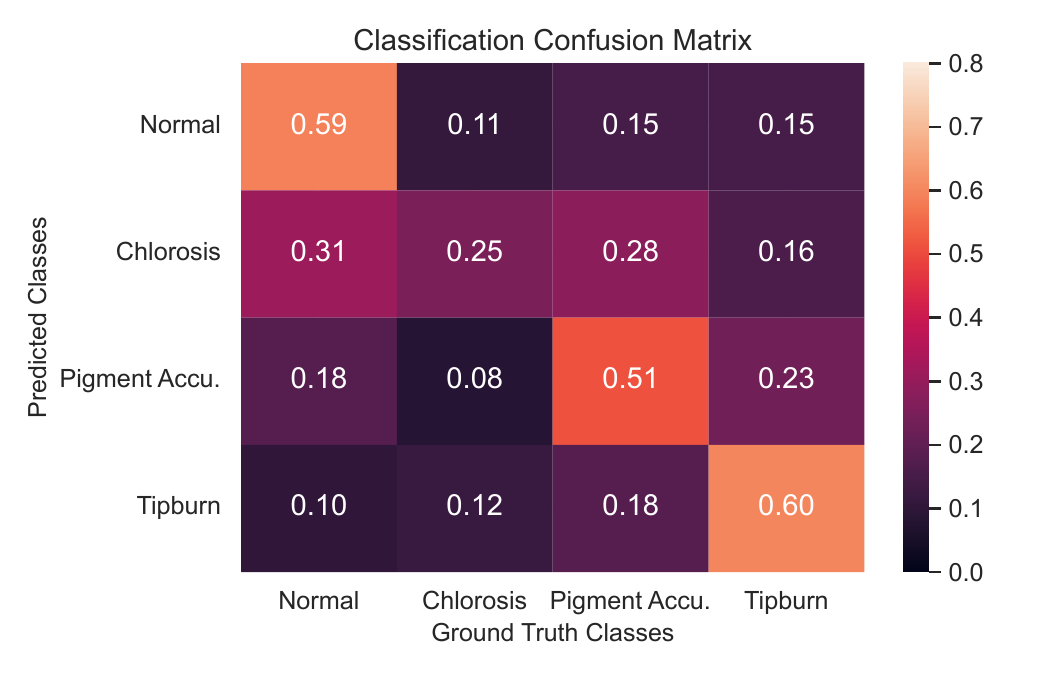}
\includegraphics[width=0.495\linewidth]{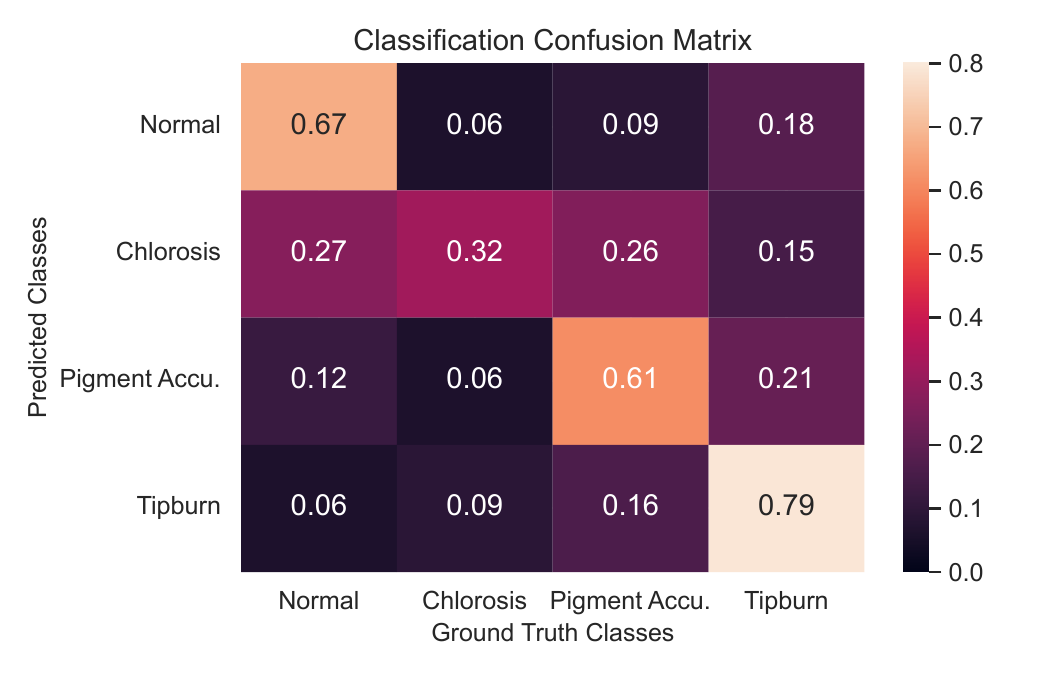}
\caption{Confusion matrices comparing the baseline YOLOv5 (left) and our proposed multi-spectral model (right). Columns denote ground truth labels, and rows denote predictions.}
\label{fig:confusion}
\end{figure}

\textbf{Confusion matrix analysis.}
Fig.~\ref{fig:confusion} presents the confusion matrices for the four annotated classes: Normal, Chlorosis, Pigment Accumulation, and Tipburn. The left panel corresponds to the baseline YOLOv5 model trained on RGB images, while the right panel shows results from our proposed multi-spectral segmentation model with transformer-based attention head.

The confusion matrices reveal several key insights regarding class-wise performance. The proposed model achieves a notable increase in segmentation accuracy across all four classes, especially for Tipburn and Normal tissue, where the correct prediction rates exceed 0.79 and 0.67, respectively. These improvements can be attributed to the enhanced spectral-spatial feature encoding provided by the transformer head and multi-channel input.

For the Tipburn class, the distinctive blackened, necrotic leaf edges provide strong visual cues. These symptoms manifest with high contrast against healthy tissue and are thus easier to localize. Moreover, the multi-spectral reflectance sharply distinguishes necrotic tissue, leading to improved segmentation performance.

The Normal class also achieves high precision due to its large area coverage and consistent appearance. Healthy tissues are characterized by uniform green coloration with regular morphology, allowing the model to learn robust representations for this class.

In contrast, the Chlorosis and Pigment Accumulation classes show more inter-class confusion, though our model significantly improves upon the baseline. Chlorosis exhibits gradual yellowing or paling of the leaf tissue, often overlapping spatially and spectrally with normal tissue. Chlorosis involves a gradual loss of chlorophyll, leading to small and gradual color shifts from green to light green to yellow, which can cause inaccurate pixel classifications, especially in early stages. This makes it inherently harder for both humans and models to distinguish from normal leaf areas, particularly if the chlorotic symptoms are mild or not spatially distinct. This contrasts to other two classes; necrosis that tends to clear and distinct boundaries and sharp color changes to black and pigment accumulation leading to vivid reddish colors with clearer segmentation in leaves, allowing them more separable in both spatial and spectral domains. 

The baseline model frequently mislabels chlorosis as normal or pigment accumulation, yielding only 0.25 accuracy. Our multi-spectral model improves this to 0.32 by capturing subtle changes in reflectance at longer wavelengths (e.g., NIR), where chlorotic tissue shows reduced reflectance.

Pigment Accumulation, typically manifested as deep purple or reddish patches, especially near the midrib or margins, poses another challenge due to its scattered and sometimes ambiguous distribution. The baseline model underperforms due to limited spectral sensitivity in RGB space. Our model increases accuracy to 0.61 by leveraging key wavelength bands that emphasize anthocyanin absorption patterns, enhancing separability from both normal and chlorotic regions.

Overall, these results in the confusion matrices  validate the benefit of using multi-spectral imaging and attention-based segmentation. Not only does the model improve correct classification rates, but it also reduces false positives across adjacent symptom categories—indicating better boundary detection and contextual understanding.

\textbf{Quantitative metric comparison.}
Table~\ref{tab:metrics} summarizes the per-class segmentation performance in terms of Intersection-over-Union (IoU) and Dice score (F1-score). The proposed multi-spectral YOLOv5 model with transformer attention head achieves consistent improvements over the baseline YOLOv5 trained on RGB images. In particular, the proposed model improves the mean IoU by $11\%$ and Dice score by $12\%$. Difficult-to-identify classes like chlorosis and pigment accumulation benefit substantially from the enhanced spectral-spatial representation.

\begin{table}[htbp]
\centering
\caption{Segmentation performance comparison between baseline YOLOv5 and proposed method (IoU and Dice scores).}
\label{tab:metrics}
\small
\begin{tabular}{|l|c|c|c|c|}
\hline
 & \multicolumn{2}{c|}{\textbf{IoU}} & \multicolumn{2}{c|}{\textbf{Dice score}} \\
\cline{2-5}
Class & Baseline YOLOv5 & Proposed Model & Baseline & Proposed \\
\hline
Normal & 0.58 & 0.67 & 0.63 & 0.74 \\
Chlorosis & 0.25 & 0.35 & 0.31 & 0.44 \\
Pigment Accum. & 0.51 & 0.61 & 0.56 & 0.68 \\
Tipburn & 0.68 & 0.79 & 0.70 & 0.82 \\
\hline
\textbf{Mean} & \textbf{0.50} & \textbf{0.61} & \textbf{0.55} & \textbf{0.67} \\
\hline
\end{tabular}
\end{table}

\begin{figure}[htbp]
\centering
\begin{subfigure}{0.325\linewidth}
    \centering
    \includegraphics[width=\linewidth]{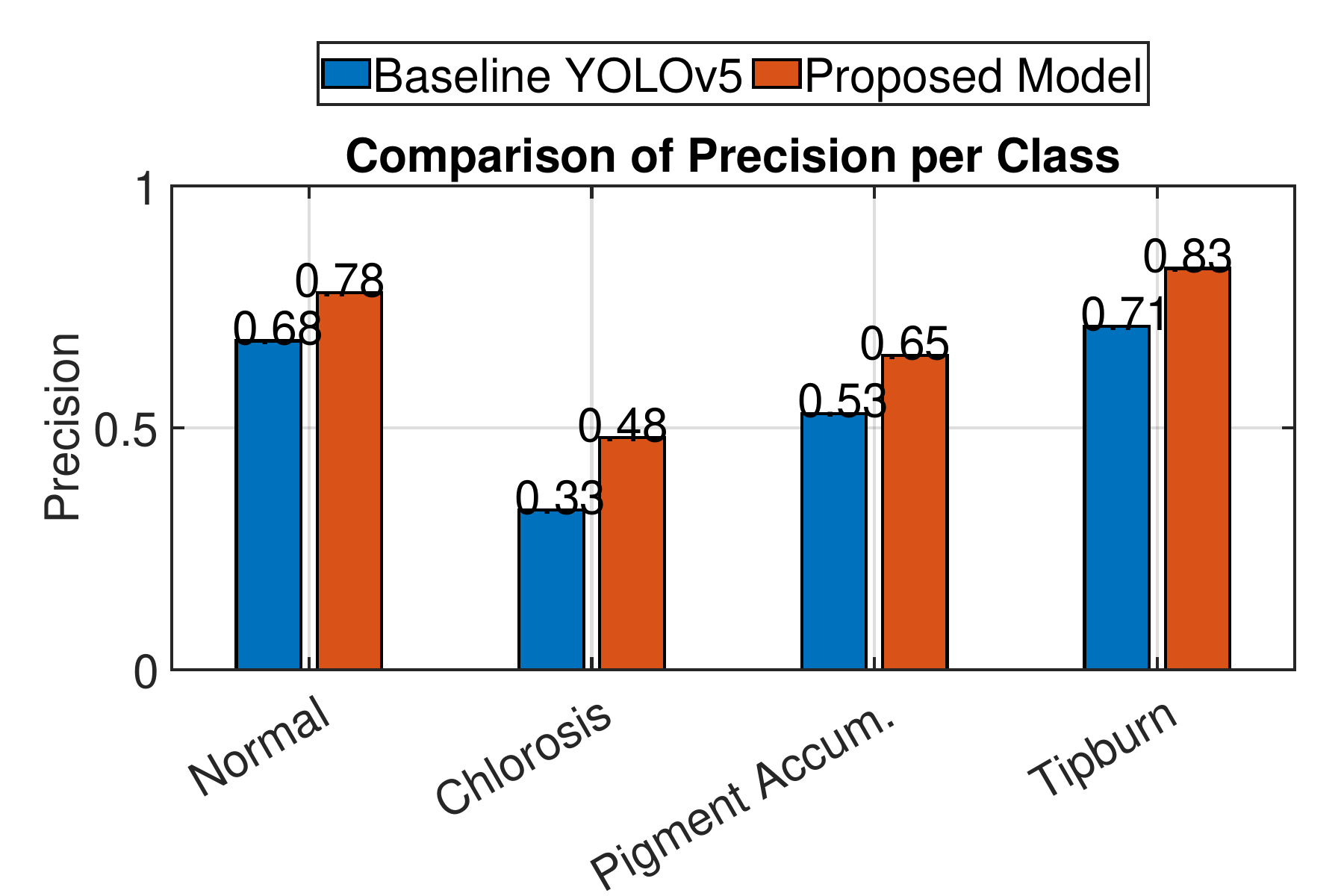}
    \caption{\centering Precision}
    \label{fig:metric_precision}
\end{subfigure}
\hfill
\begin{subfigure}{0.325\linewidth}
    \centering
    \includegraphics[width=\linewidth]{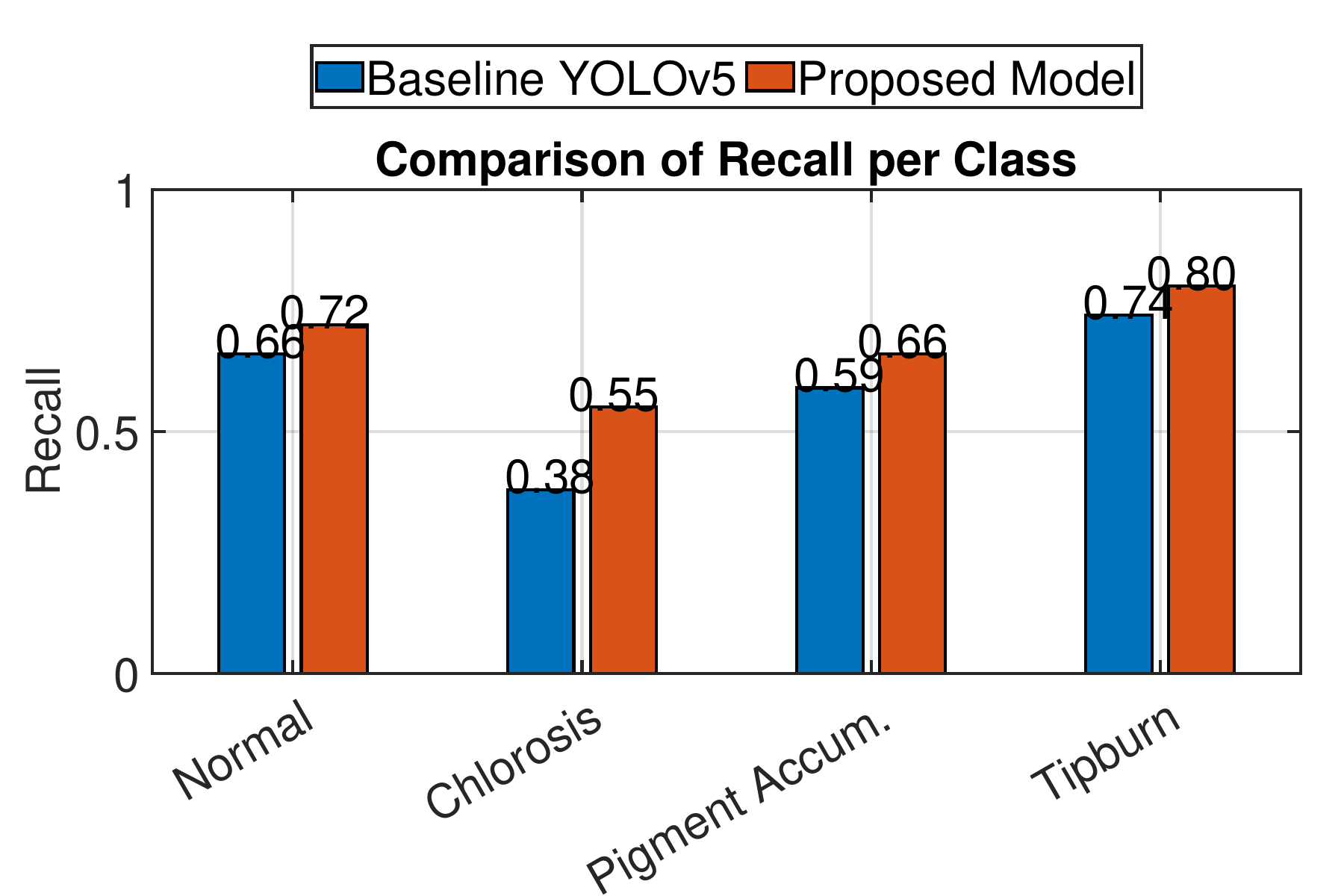}
    \caption{\centering Recall}
    \label{fig:metric_recall}
\end{subfigure}
\hfill
\begin{subfigure}{0.325\linewidth}
    \centering
    \includegraphics[width=\linewidth]{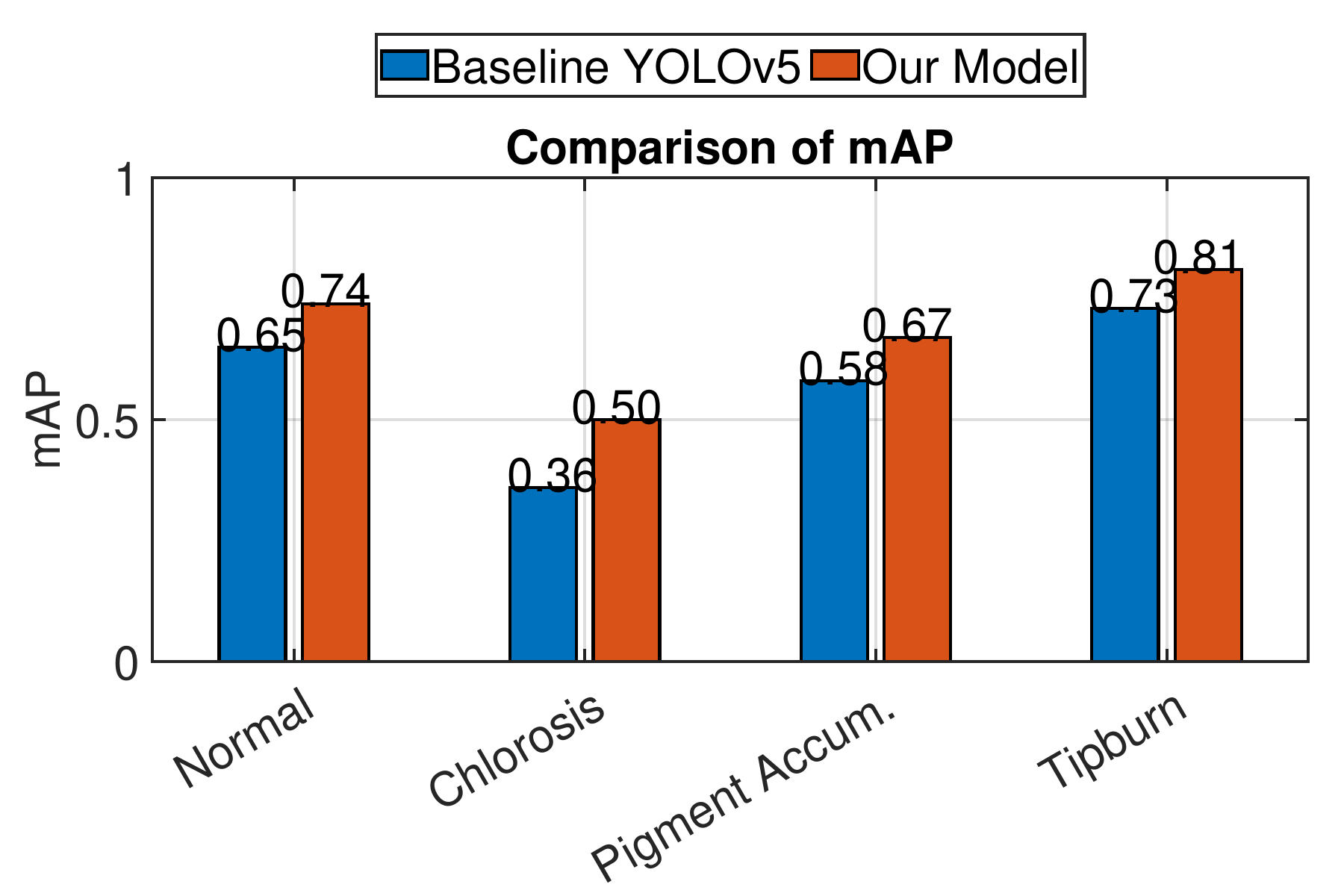}
    \caption{\centering mAP}
    \label{fig:metric_map}
\end{subfigure}
\caption{Comparison of Precision, Recall, and mAP across all classes for baseline YOLOv5 and the proposed model.}
\label{fig:metric_all}
\end{figure}

To provide a more comprehensive evaluation, we also compare the models in terms of class-wise Precision, Recall, and mean Average Precision (mAP) scores, as shown in Fig.~\ref{fig:metric_precision}, Fig.~\ref{fig:metric_recall}, and Fig.~\ref{fig:metric_map} respectively. These metrics further validate the superior performance of our proposed model over the baseline YOLOv5.

From these plots, it is evident that our model achieves substantial gains across all evaluation metrics. For example, the Precision for Chlorosis improves from $0.33$ to $0.48$, and Recall increases from $0.38$ to $0.55$. Similarly, the mAP for Chlorosis improves from $0.36$ to $0.50$, showing that the proposed model is more robust at correctly identifying subtle and overlapping symptoms. For well-defined classes like Tipburn and Normal, both Precision and Recall exceed $0.80$, which confirms the model’s capability to capture strongly localized symptoms with distinct boundaries.

These improvements can be attributed to the model’s ability to process all nine spectral channels, enabling it to detect variations in leaf pigment, reflectance, and morphology more effectively than standard RGB-only approaches. The use of a transformer-based attention head further enhances contextual learning and boundary refinement, especially for small or sparsely distributed stress symptoms.

\begin{figure}[htbp]
\centering
\begin{subfigure}{0.32\linewidth}
    \centering
    \includegraphics[width=\linewidth]{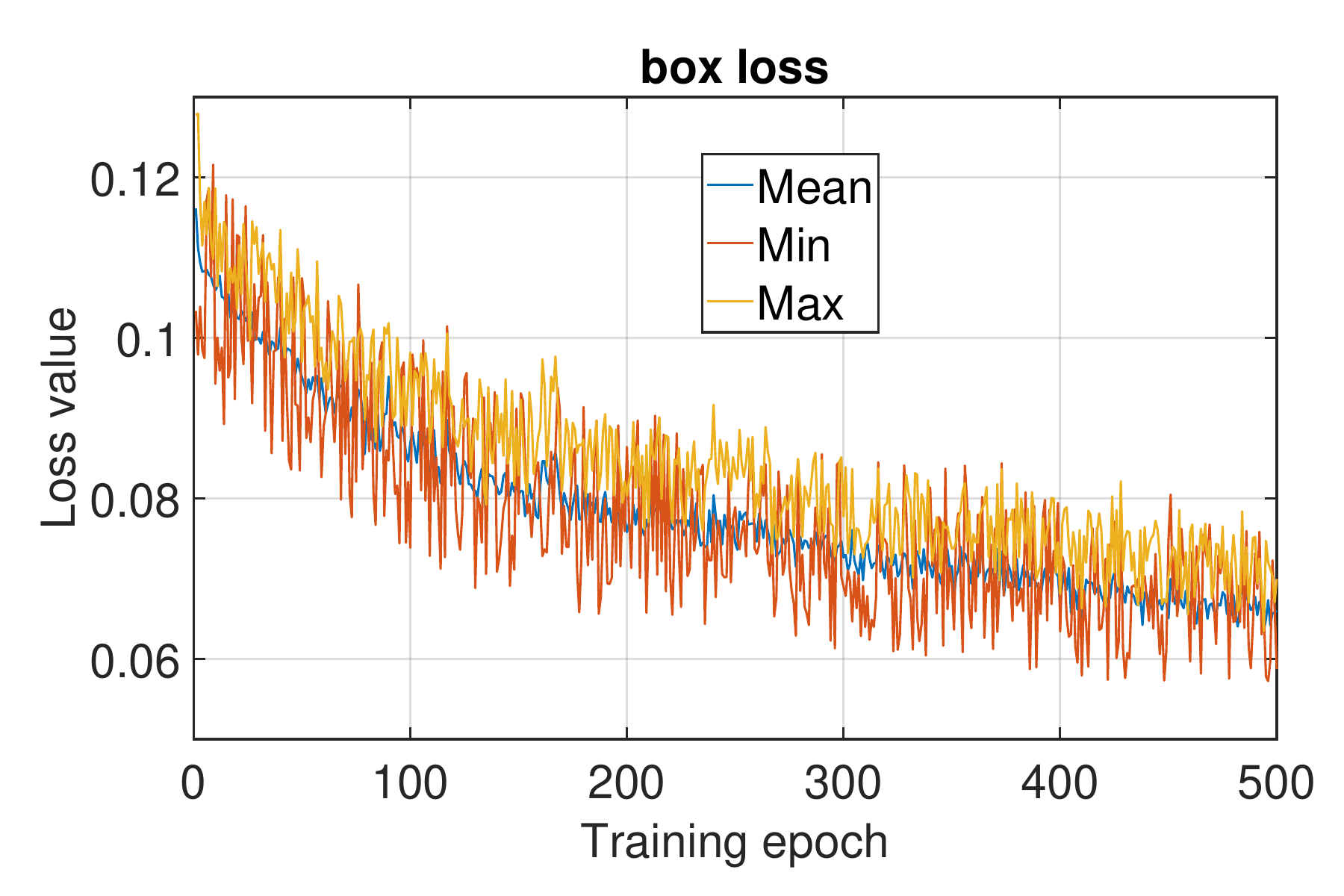}
    \caption{\centering box loss}
    \label{fig:metric_recall}
\end{subfigure}
\begin{subfigure}{0.32\linewidth}
    \centering
    \includegraphics[width=\linewidth]{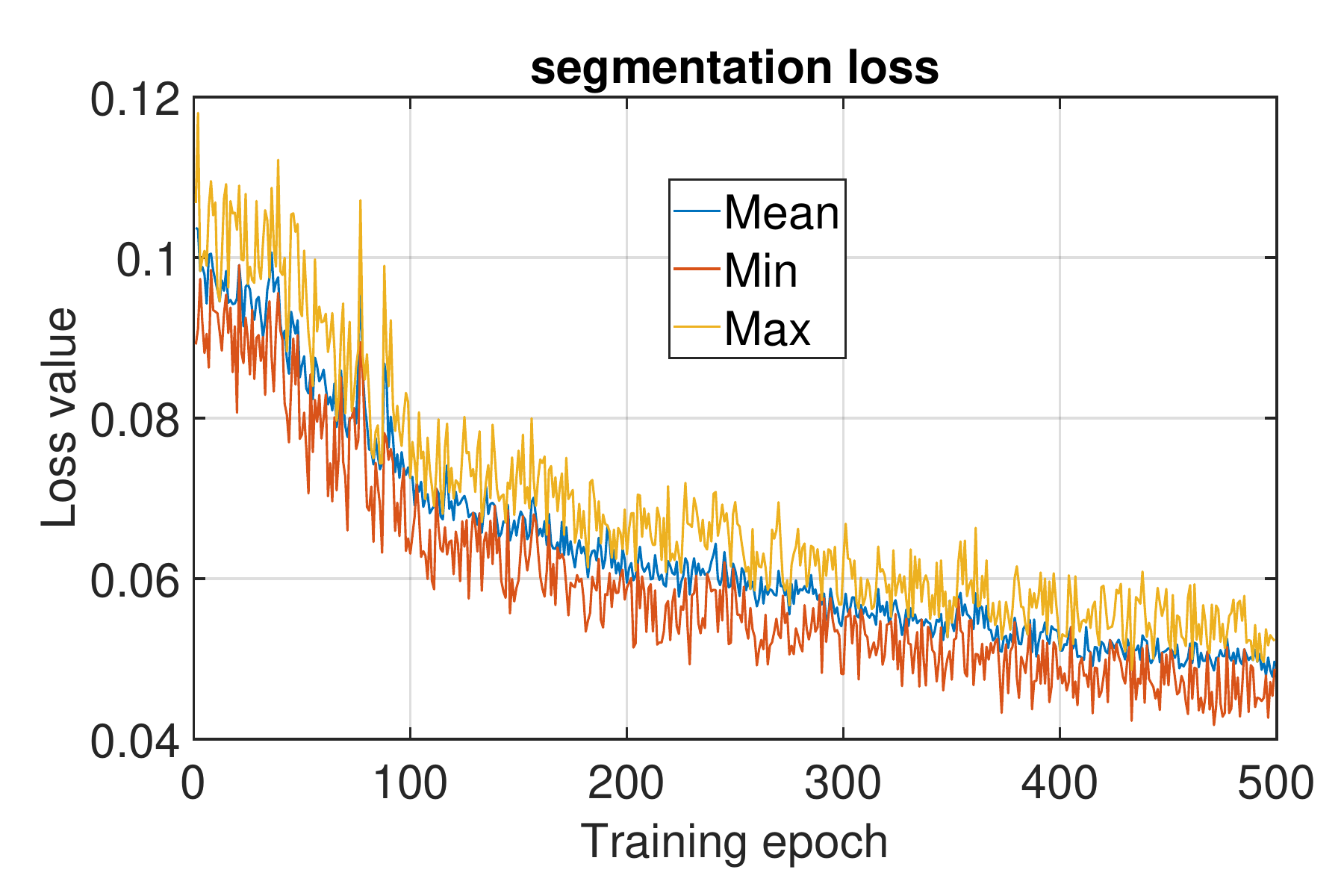}
    \caption{\centering segmentation loss}
    \label{fig:metric_recall}
\end{subfigure}
\begin{subfigure}{0.32\linewidth}
    \centering
    \includegraphics[width=\linewidth]{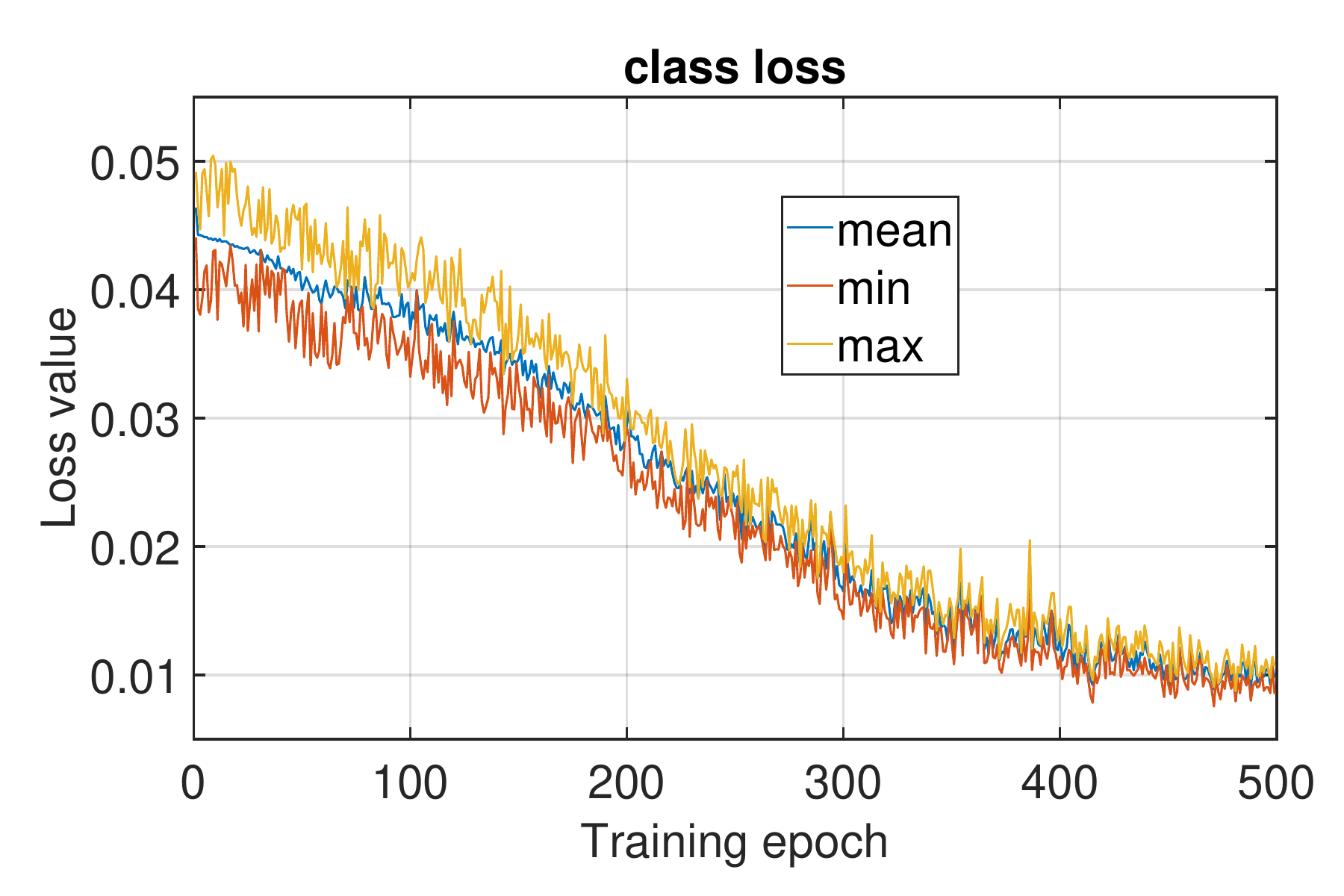}
    \caption{\centering class loss}
    \label{fig:metric_recall}
\end{subfigure}\\
\begin{subfigure}{0.32\linewidth}
    \centering
    \includegraphics[width=\linewidth]{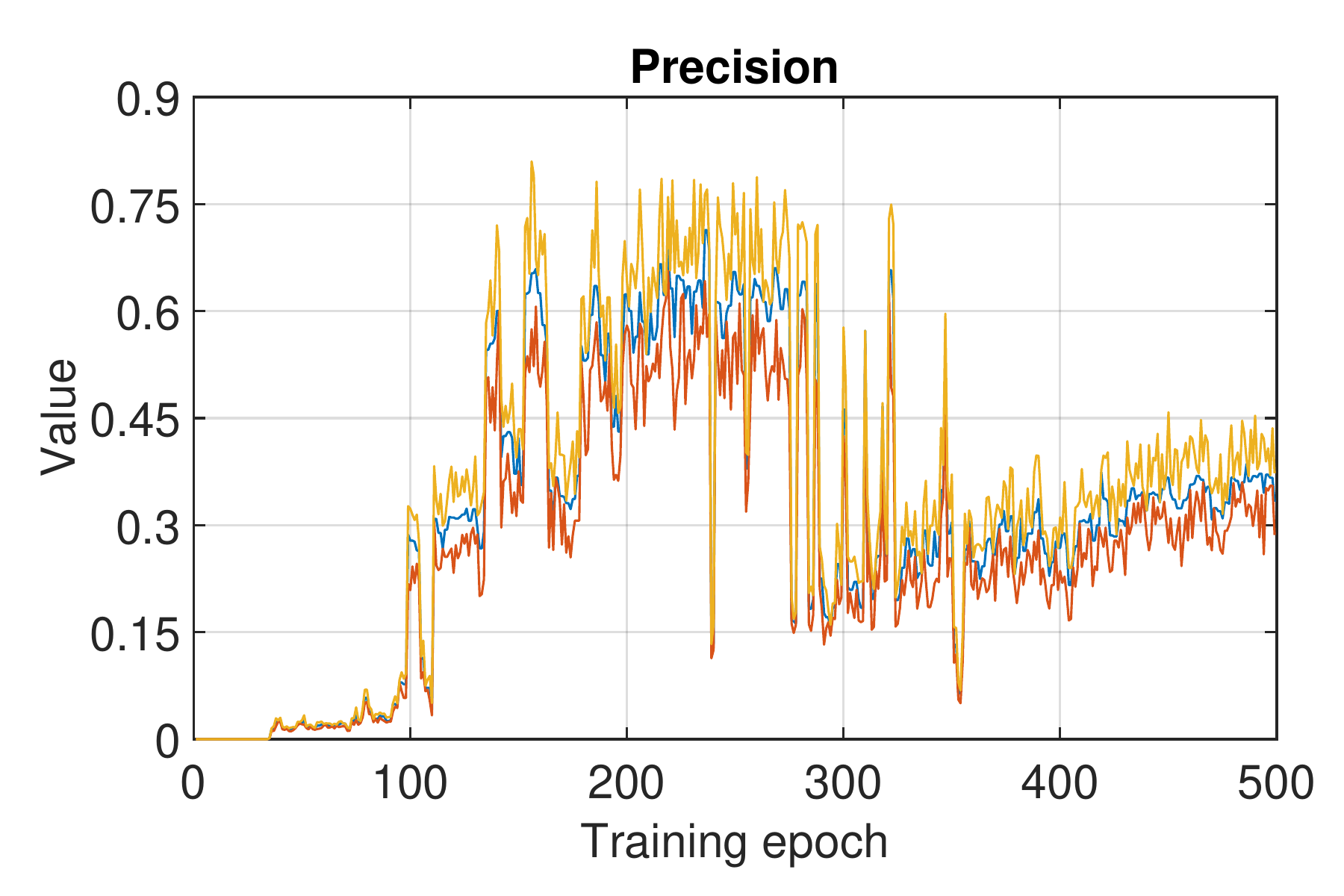}
    \caption{\centering Precision}
    \label{fig:metric_recall}
\end{subfigure}
\begin{subfigure}{0.32\linewidth}
    \centering
    \includegraphics[width=\linewidth]{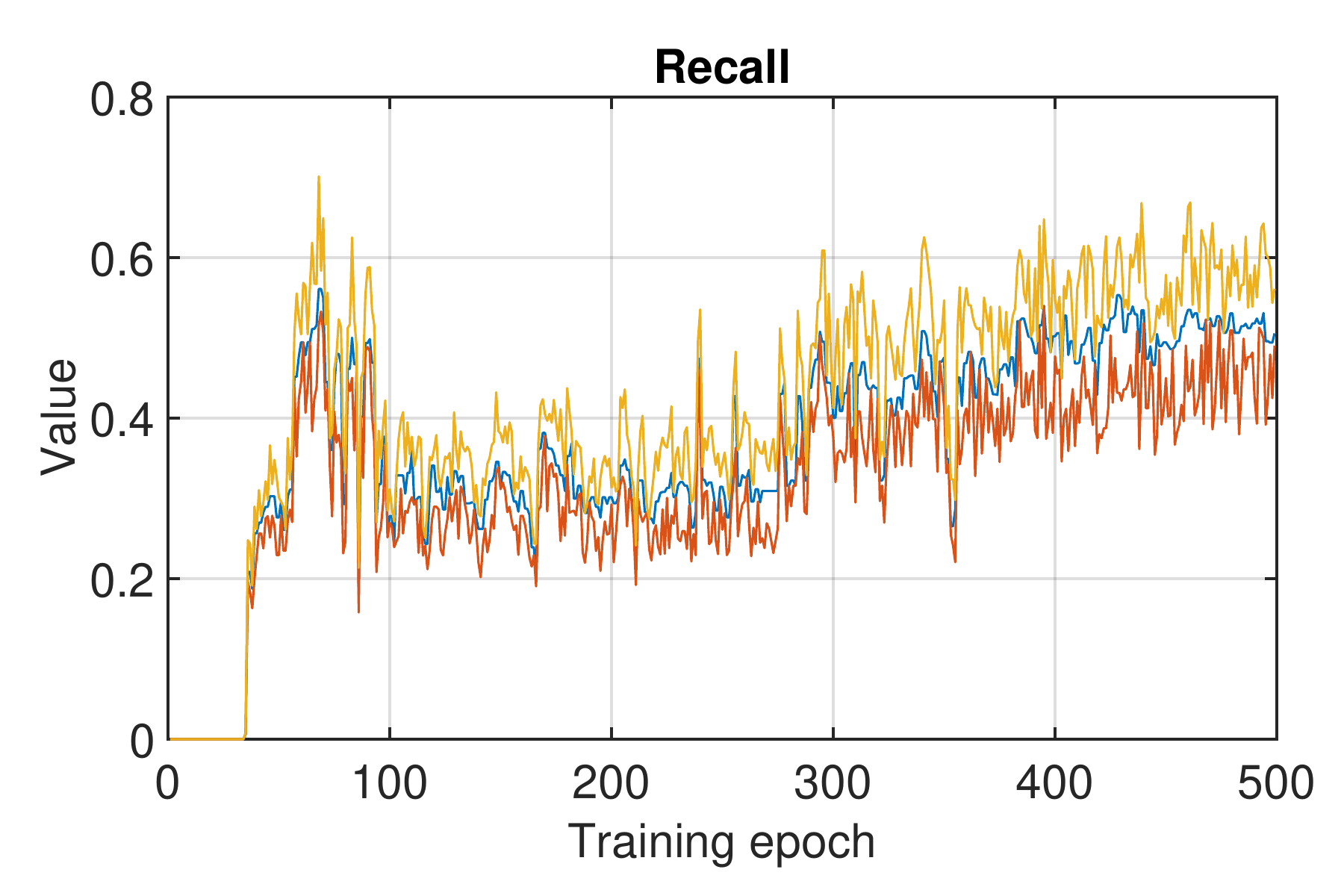}
    \caption{\centering Recall}
    \label{fig:metric_recall}
\end{subfigure}
\begin{subfigure}{0.32\linewidth}
    \centering
    \includegraphics[width=\linewidth]{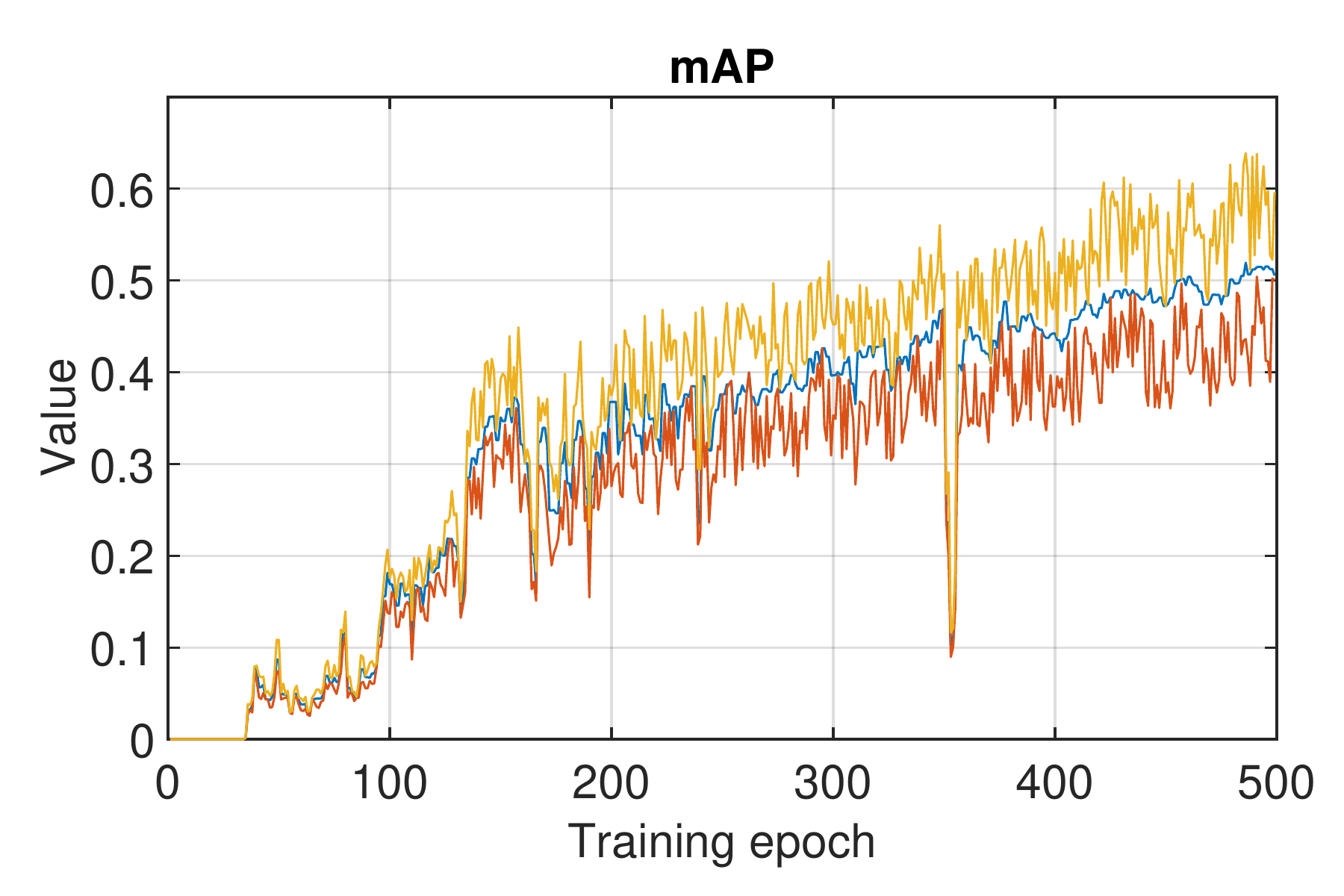}
    \caption{\centering mAP}
    \label{fig:metric_recall}
\end{subfigure}
\caption{Training profile of the proposed model on the multi-spectral test dataset.}
\label{fig:train_accu}
\end{figure}
\textbf{Training profile and convergence.}
The training profile of the proposed model over 500 epochs is shown in Fig.~\ref{fig:train_accu}. The top row illustrates loss values for box localization, segmentation mask, and classification. All three loss components converge steadily, with segmentation loss showing a distinct "warm-up" behavior in early epochs followed by rapid decline.

The second row in Fig.~\ref{fig:train_accu} depicts precision, recall, and mAP over epochs. Precision and recall values improve concurrently, indicating the model learns robust class boundaries. The mAP curve stabilizes above 0.70, reflecting strong overall performance across confidence thresholds.

\begin{figure}[htbp]
\centering
\includegraphics[width=0.9\linewidth]{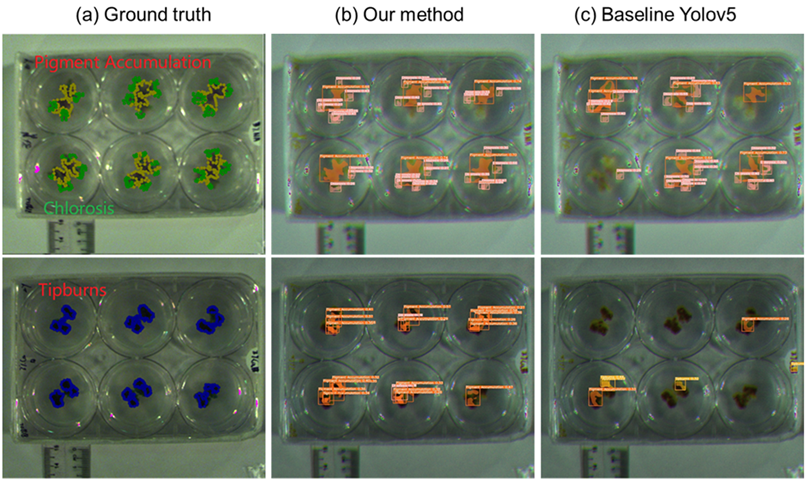}
\caption{Visual comparison of segmentation results: Ground truth (left), our proposed method prediction (middle), and baseline YOLOv5 prediction (right).}
\label{fig:vis}
\end{figure}

\textbf{Visual segmentation comparison.}
To evaluate qualitative differences, Fig.~\ref{fig:vis} presents typical segmentation results on chlorosis and pigment accumulation samples. Each group contains three images: the ground truth annotation, our proposed model's prediction, and the baseline YOLOv5 prediction.

The proposed method captures finer boundaries and more complete lesion regions, particularly in cases of scattered chlorosis and marginal pigment accumulation. The baseline YOLOv5 model, by contrast, misses faint lesions or merges adjacent regions, indicating limited spectral sensitivity.

Overall, the results confirm that incorporating multi-spectral channels and a transformer-based attention head significantly improves segmentation accuracy, especially for small or subtle plant stress symptoms. This demonstrates the value of combining spectral and spatial cues in plant phenotyping tasks.

\section{Conclusion}
This study presents a multi-spectral plant health analysis framework based on an enhanced YOLOv5 segmentation model with a transformer-based attention head. By adapting the model architecture to accept nine-channel multi-spectral input and incorporating self-attention mechanisms, we improve the model’s ability to detect and localize subtle and spatially distributed symptoms such as chlorosis, pigment accumulation, and tipburn.

Through a series of controlled nutrient-deficiency experiments and pixel-wise annotated datasets, we demonstrate that the proposed method significantly outperforms the baseline RGB-only YOLOv5 model. Quantitative results show consistent improvements across IoU and Dice metrics, while qualitative visualizations confirm better boundary delineation and symptom sensitivity.

This work highlights the advantages of combining spectral and spatial features for precision plant phenotyping, especially in scenarios where early symptom detection is critical.

\textbf{Future Work.} While the current model shows strong performance, future research may explore the following directions: i) Integration of temporal information from time-series multi-spectral images to improve early detection and progression tracking; ii) Lightweight model design via channel pruning or knowledge distillation for deployment on edge devices in greenhouses or farms. iii) Automatic band selection or spectral attention modules to identify and prioritize the most informative wavelengths. iv) Expansion of the dataset to include more crop types, field conditions, and stress scenarios for generalizability. 

From a field deployment perspective, the proposed model offers clear potential for adaptation to real-world agricultural settings. The compact design of modern multi-spectral sensors allows for easy integration into drone-based or handheld platforms, facilitating in-situ crop monitoring. Furthermore, the lightweight architecture of the YOLO-based model supports efficient inference on edge devices, enabling real-time stress detection in the field. For robust performance, future deployment pipelines should incorporate domain adaptation techniques to handle environmental variations such as lighting, background clutter, and leaf occlusions.
\bibliographystyle{elsarticle-num}
\bibliography{sample}

\end{document}